\documentclass[a4paper, 10pt, conference]{ieeeconf}      

\IEEEoverridecommandlockouts                              

\usepackage{tikz}
\usepackage{pgfplots}
\usepackage{amsmath} 
\usepackage{amsfonts}
\usepackage{booktabs}
\usepackage{subcaption}
\usepackage{flushend}
\usepackage{cite}
\makeatletter
\let\NAT@parse\undefined
\makeatother
\usepackage[bookmarks=true]{hyperref}

\tikzset{
every node/.style = {text height=0.8em,text depth=.15em}
}

\pgfplotsset{compat=1.10}

\usetikzlibrary{shapes}
\usetikzlibrary{arrows,arrows.meta}
\usetikzlibrary{fit}
\usetikzlibrary{calc}
\usetikzlibrary{positioning}

\title{\LARGE \bf
NeuralMeshing: Complete Object Mesh Extraction from Casual Captures
}

\author{Floris Erich*$^{1}$, Naoya Chiba$^{2}$, Abdullah Mustafa$^{1}$, Ryo Hanai$^{1}$, \\Noriaki Ando$^{1}$, Yusuke Yoshiyasu$^{1}$ and Yukiyasu Domae$^{1}$
\thanks{$^{*}$Corresponding author, reachable at floris.erich@aist.go.jp}
\thanks{$^{1}$Floris Erich, Abdullah Mustafa, Ryo Hanai, Noriaki Ando, Yusuke Yoshiyasu and Yukiyasu Domae are with the National Institute of Advanced Industrial Science and Technology (AIST), Japan.}%
\thanks{$^{2}$Naoya Chiba is with University of Osaka.}%
}

\begin{document}

\maketitle
\thispagestyle{empty}
\pagestyle{empty}

\begin{figure*}
  \centering
  \includegraphics[width=0.95\textwidth]{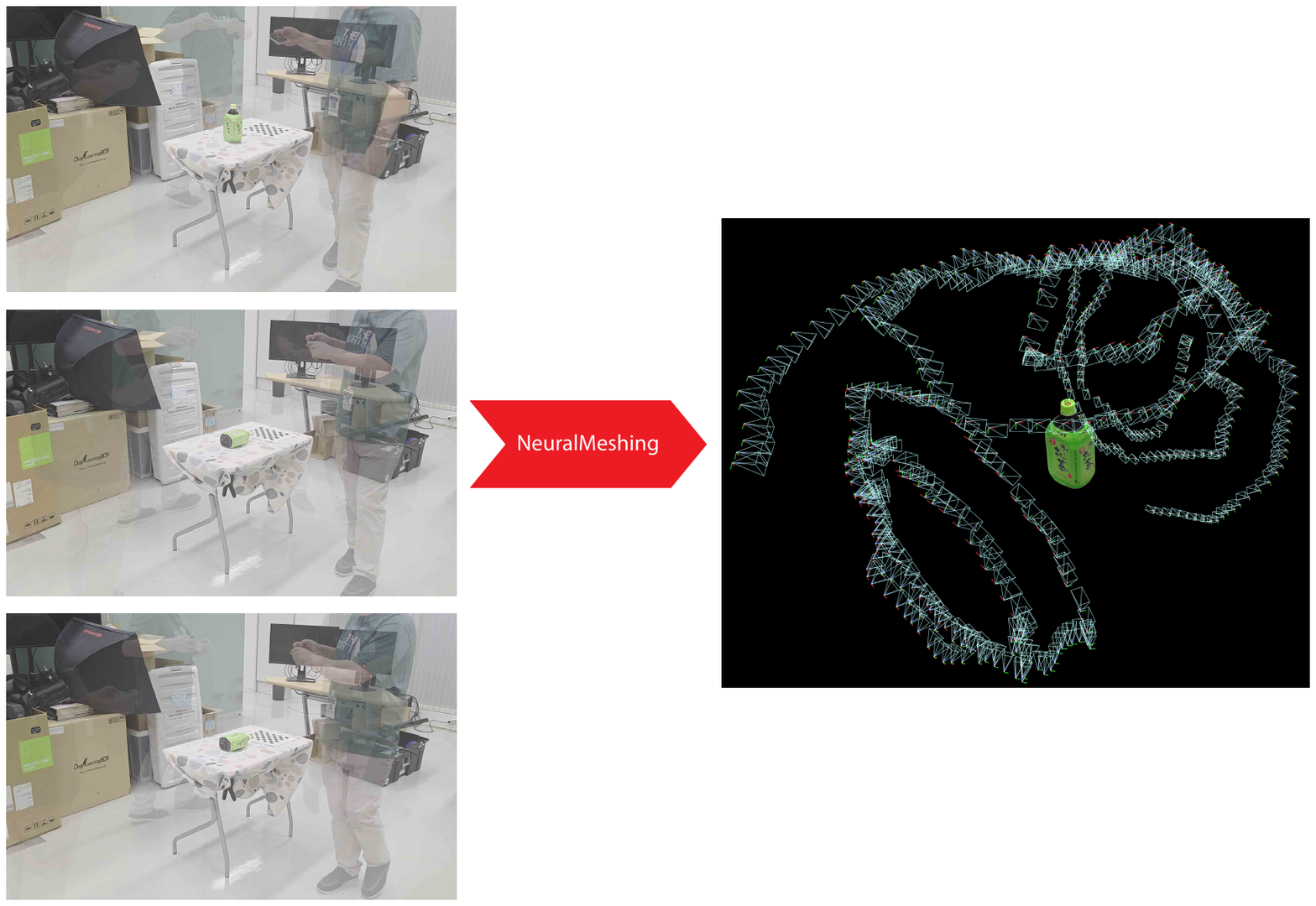}
  \caption{Example of a neural reconstruction of a partially translucent object based on three videos (object standing upright, object placed on its back side and object placed on its front side).}
\end{figure*}

\begin{abstract}

  How can we extract complete geometric models of objects that we encounter in our daily life, without having access to commercial 3D scanners?
  In this paper we present an automated system for generating geometric models of objects from two or more videos.
  Our system requires the specification of one known point in at least one frame of each video, which can be automatically determined using a fiducial marker such as a checkerboard or Augmented Reality (AR) marker.
  The remaining frames are automatically positioned in world space by using Structure-from-Motion techniques.
  By using multiple videos and merging results, a complete object mesh can be generated, without having to rely on hole filling.
  Code for our system is available from \url{https://github.com/FlorisE/NeuralMeshing}.

\end{abstract}

\section{INTRODUCTION}

\begin{figure*}
\centering
  \includegraphics[width=\textwidth]{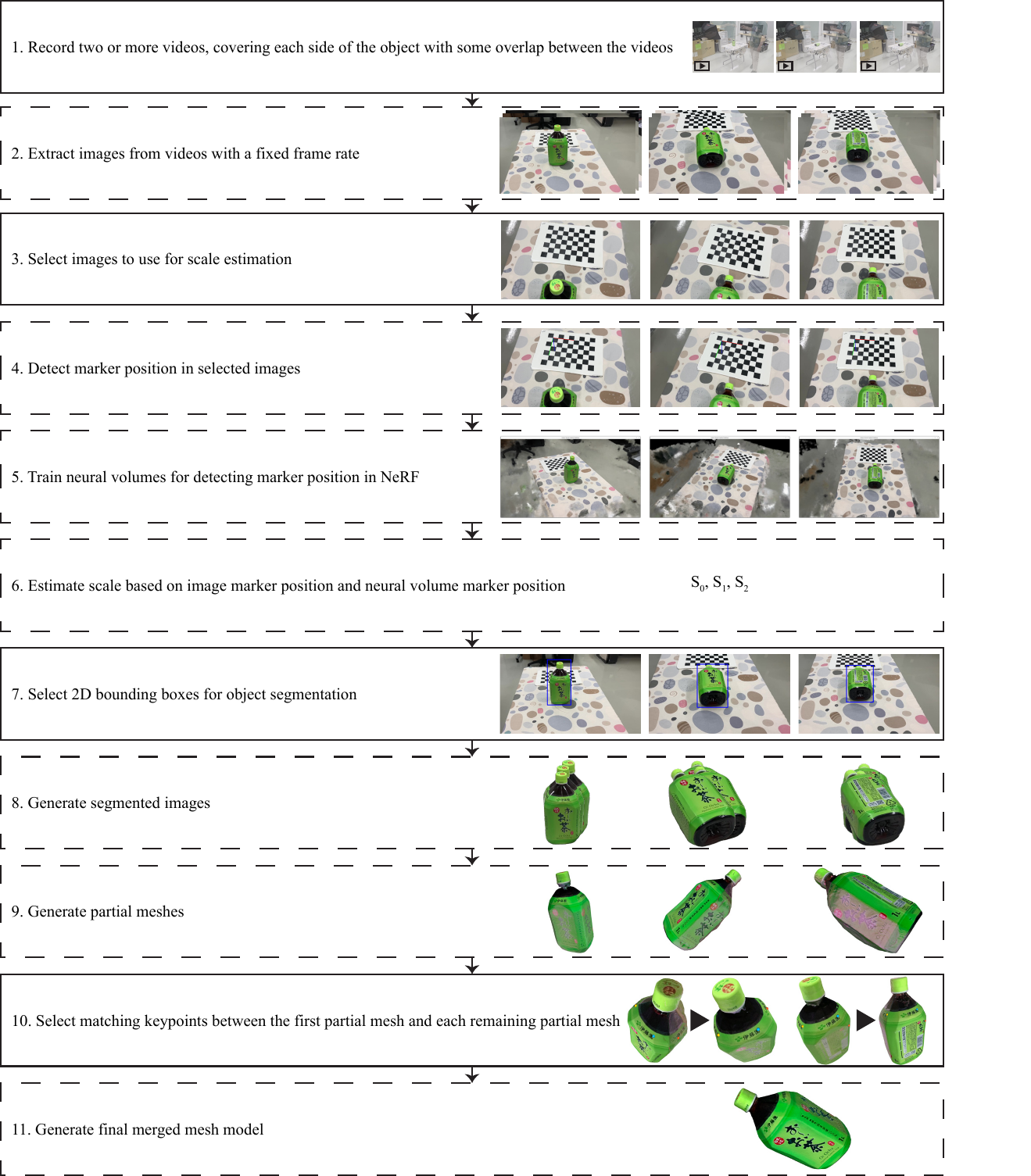}
\caption{Flowchart of steps executed to generate a mesh. Manual steps have a solid line while completely automated steps have a dashed line. Zoom in for more details.}
\label{fig:flowchart}
\end{figure*}

Triangle meshes are used in various fields such as computer vision, computer graphics and robotics.
They can be obtained through manual 3D modeling or through automation, known as 3D scanning.
3D scanning equipment is expensive, with prices ranging from thousands to tens of thousands of dollars.
Open source software such as COLMAP and GLOMAP offer a free alternative, but they are complicated to use when the goal is to scan objects instead of environments.

We present a system that integrates various state-of-the-art techniques into a user friendly pipeline for generating a complete object mesh with realistic proportions.
Our system only requires a video camera (e.g.\ smartphone or USB camera) and a surface to place the object and fiducial marker on.
Ficudial markers can be printed on paper and are thus easy to create.

Our method allows for generating complete object models of objects with a diffuse texture.
The method can also be applied on reflective or transparent objects, with some degree of success.
Merging of multiple videos is achieved by the operator manually selecting matching points, however this requires that the object has at least three easily identifiable points shared between videos.
Such points have unique geometry or colors allowing the operator to easily identify them in multiple videos.

We improve upon state-of-the-art methods such as NeuS2~\cite{neus2} by integrating it in a pipeline for complete mesh generation from video, as well as improving the texture quality.
NeuS2 can automatically construct a neural representation from video data by integrating COLMAP and can export meshes, but the meshes will be unscaled and NeuS2 cannot merge multiple videos of an object.
The texture quality of meshes exported using NeuS2 is often poor due to using an inaccurate camera direction when estimating mesh colors.
By sampling from every direction, the logical direction vector for estimating the color for a vertex is opposite of the surface normal vector.

\section{BACKGROUND}

3D reconstruction from images is a core topic of the computer vision field, with various methods proposed to create mesh representations~\cite{schoenberger2016mvs, schoenberger2016sfm} or novel view renderings~\cite{mildenhall2021}.
Such techniques can be integrated with hardware systems such as rotation stages~\cite{erich2022}, camera rigs~\cite{erich2023neuralscanning} or robots~\cite{qiu2025}, however the cost of such equipment is excessive for non-commercial applications.
Smartphone-based scanning applications have been released, but these have limitations such as requiring the smartphone to be equipped with a depth sensor or rely on AI techniques that do not generalize well to novel objects.
Extremely sparse view mesh reconstruction generative AI methods have been released~\cite{xiang2025, shi2024b, Liu2023Zero1to3ZO}, but these do not reproduce the same amount of detail (such as readable text) as we aim for in this research.
Dataset annotation tools such as HANDAL~\cite{guo2023} or NeuralLabeling~\cite{erich2024neurallabeling-iros} also generate meshes, however these are often untextured or have incomplete geometry.

We present a cost-effective method for creating meshes by integrating marker-based scale estimation and state-of-the-art mesh generation techniques.

\section{METHOD}
An overview of each step in our method is shown in Fig.~\ref{fig:flowchart}.
Object meshes are defined as sets of vertices $V \in \mathbb R^3$, edges $E = V \times V$, faces $F = V \times V \times V$ and vertex colors $C \in \mathbb R^3$.
Our method takes as input videos $X \in \{X_0, \ldots, X_N\}$ consisting of frames $I_X \in \mathbb R^{H \times W \times 3 \times F}$, where $W$, $H$, $F$ are the frame width, frame height and number of frames, respectively.
A frame rate is specified by the operator during recording, such that there is enough overlap between frames, but not too many frames to significantly affect the processing time or to exhaust computational resources such as GPU memory.
During recording, the operator moves the camera around the object carefully to avoid motion blur while capturing all the relevant sides of the object in its current position, while keeping the object stationary.
Recording is repeated to capture sides that were occluded in other videos.
For objects that can be easily flipped, two recordings suffice.
For objects that cannot easily be flipped, it might be necessary to capture 3 videos or more of the object.

We use Structure-from-Motion (SfM) algorithms~\cite{schoenberger2016sfm, schoenberger2016mvs} to estimate the camera instrinsics $C_X$ and extrinsics $T_X$ for all $I_X$, both being relative to a yet unknown scale factor $S_X$.
We assume camera intrinsics are static within a video and enforce this in the SfM algorithms by detecting a single camera model for each video.
We determine and store camera intrinsics using the OpenCV camera model, consisting of 9 parameters: $k_1, k_2, k_3, p_1, p_2, f_x, f_y, c_x, c_y$.
We store camera extrinsics as a transformation matrix $T \in \mathbb R^{4 \times 4}$ consisting of a rotation matrix \verb|T[:3,:3] = R|, a translation vector \verb|T[:3, 3] = t| and \verb|T[3, :] = [0, 0, 0, 1]|.

To segment the object from its background we use Segment Anything 2 (SAM2)~\cite{kirillov2023a, ravi2025} in bounding-box mode.
The operator draws a 2D bounding box around the target object, after which the object is tracked through all video frames and produces $M \in \mathbb Z^{H \times W \times F}$.
Broadcasting combines $I_X$ and $M$ to generate masked images $I_{X,M} \in \mathbb R^{H \times W \times 3 \times F}$.

We construct a neural field~\cite{mildenhall2021} by inputting $I_{X,M}$, $C_X$ and $T_X$ using \emph{instant-ngp}~\cite{muller2022}, which allows for fast training by using a multi-level hash encoding.
A surface loss term is used in order to produce a neural field in which the density of the neural field matches the surface of the object~\cite{neus2}.
We then extract a partial object mesh $M_X$ by applying the marching cubes (MC) algorithm~\cite{lorensen1987} on the neural field using a resolution of height $H_\text{MC}$, width $W_\text{MC}$ and depth $Z_\text{MC}$.
We use an implementation of NeuS2 that uses opposite surface normals as MLP query direction for determining vertex color.

For each video $X$, the operator selects in at least one frame one point for which the ground truth position $P_X \in \mathbb R^3$ is known.
In practice, we place a marker and detect its position, and the operator only has to select which frames to use for marker detection.
To determine the ground truth position, we use marker detection with the known marker properties such as radius for a circle or square side length for a checkerboard.
To determine the position $P_{X, \text{NeRF}} \in \mathbb R^3$ of the marker in the reconstruction, which is subject to a scale factor, we render the neural volume with the determined camera intrinsics and use the extrinsics of the manually selected frames by the operator.
We can now calculate the scale factor $S_X = \|P_X\| / \|P_{X, \text{NeRF}}\|$.
In case the operator selected multiple frames to use for scaling, we simply repeat the scaling procedures for each selected frame and take the mean of the individually determined scaling factors.

We scale the partial object meshes $M_{X, 0}, \ldots, M_{X, N}$ to a uniform scale.
The operator then selects at least three matching points on each partial object mesh to perform coarse registration via point correspondences.
After coarse registration, the system performs fine registration using Iterative Closest Points (ICP)~\cite{zhang1994}.
This procedure results in the construction of a joint neural field.
We repeat the same mesh extraction technique discussed earlier to produce the object mesh $M$ with per-vertex coloring.
Because the object meshes will have a large file size due to a high vertex count, techniques for mesh decimation and texture extraction can be automatically applied.

\section{EVALUATION}

\begin{table*}[t]
  \addtolength{\tabcolsep}{-4pt}
\caption{Quantitatively evaluating our method compared to ablations}
\centering
\begin{tabular}{ ccccccccccccccccccc }
  \toprule
Method & \multicolumn{3}{c}{Green Tea Bottle} & \multicolumn{3}{c}{Noodles} & \multicolumn{3}{c}{Beer Can} & \multicolumn{3}{c}{Mustard Bottle} & \multicolumn{3}{c}{Coffee Ground} & \multicolumn{3}{c}{Mean} \\
 & MAE & RMSE & PSNR & MAE & RMSE & PSNR & MAE & RMSE & PSNR & MAE & RMSE & PSNR & MAE & RMSE & PSNR & MAE & RMSE & PSNR \\
\midrule
NeRF & 0.22 & 0.26 & 11.72 & 0.17 & 0.22 & 13.23 & 0.31 & 0.37 & 8.54 & 0.16 & 0.21 & 13.67 & 0.25 & 0.29 & 10.65 & 0.22 & 0.27 & 11.56 \\
NeuS2 & 0.26 & 0.30 & 10.66 & 0.17 & 0.21 & 13.77 & 0.19 & 0.24 & 12.29 & 0.18 & 0.23 & 12.79 & 0.23 & 0.29 & 11.12 & 0.20 & 0.25 & 12.13 \\
NeuralMeshing & 0.10 & 0.14 & 17.27 & 0.09 & 0.13 & 18.28 & 0.17 & 0.22 & 13.12 & 0.07 & 0.10 & 20.46 & 0.14 & 0.19 & 14.86 & 0.11 & 0.16 & 16.80 \\
\bottomrule
\end{tabular}
\label{tab:quantitative_results}
\end{table*}

\begin{figure*}
  \centering
   \begin{subfigure}[T]{0.10\textwidth}
   \includegraphics[width=0.96\textwidth]{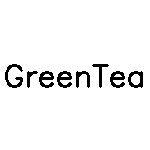}
     \vspace{1mm}
   \includegraphics[width=0.96\textwidth]{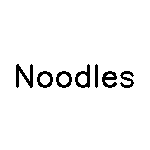}
     \vspace{1mm}
   \includegraphics[width=0.96\textwidth]{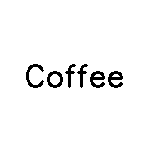}
     \vspace{1mm}
   \includegraphics[width=0.96\textwidth]{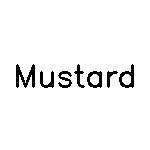}
     \vspace{1mm}
   \includegraphics[width=0.96\textwidth]{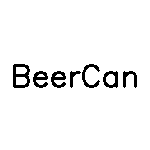}
     \vspace{1mm}
   \includegraphics[width=0.96\textwidth]{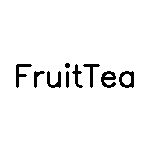}
     \caption{Labels}
   \end{subfigure}
    \begin{subfigure}[T]{0.20\textwidth}
   \includegraphics[width=0.48\textwidth]{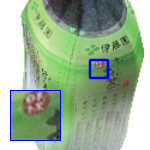}
   \includegraphics[width=0.48\textwidth]{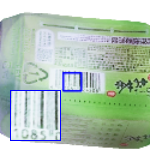}
   \vspace{1mm}
   \includegraphics[width=0.48\textwidth]{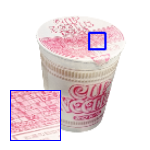}
   \includegraphics[width=0.48\textwidth]{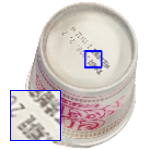}
   \vspace{1mm}
   \includegraphics[width=0.48\textwidth]{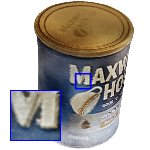}
   \includegraphics[width=0.48\textwidth]{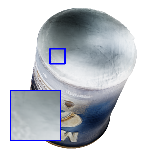}
   \vspace{1mm}
   \includegraphics[width=0.48\textwidth]{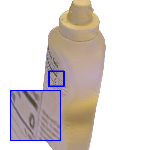}
   \includegraphics[width=0.48\textwidth]{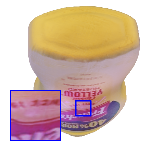}
   \vspace{1mm}
   \includegraphics[width=0.48\textwidth]{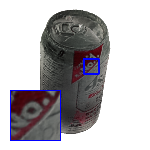}
   \includegraphics[width=0.48\textwidth]{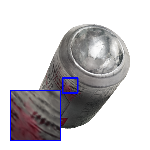}
   \vspace{1mm}
   \includegraphics[width=0.48\textwidth]{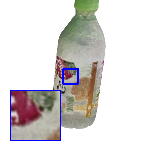}
   \includegraphics[width=0.48\textwidth]{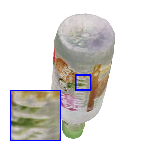}
   \caption{NeuS2}
   \end{subfigure}
   \begin{subfigure}[T]{0.20\textwidth}
   \includegraphics[width=0.48\textwidth]{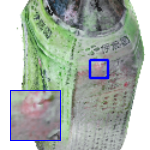}
   \includegraphics[width=0.48\textwidth]{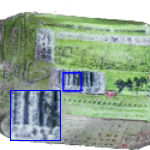}
   \vspace{1mm}
   \includegraphics[width=0.48\textwidth]{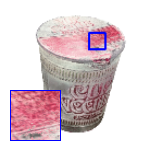}
   \includegraphics[width=0.48\textwidth]{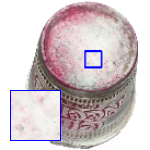}
   \vspace{1mm}
   \includegraphics[width=0.48\textwidth]{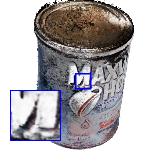}
   \includegraphics[width=0.48\textwidth]{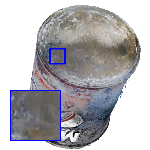}
   \vspace{1mm}
   \includegraphics[width=0.48\textwidth]{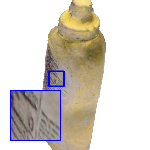}
   \includegraphics[width=0.48\textwidth]{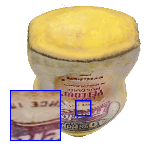}
   \vspace{1mm}
   \includegraphics[width=0.48\textwidth]{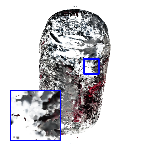}
   \includegraphics[width=0.48\textwidth]{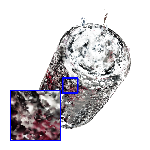}
   \vspace{1mm}
   \includegraphics[width=0.48\textwidth]{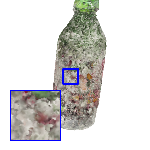}
   \includegraphics[width=0.48\textwidth]{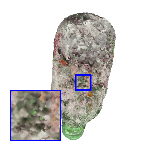}
   \caption{NeRF}
   \end{subfigure}
   \begin{subfigure}[T]{0.20\textwidth}
   \includegraphics[width=0.48\textwidth]{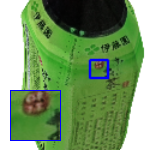}
   \includegraphics[width=0.48\textwidth]{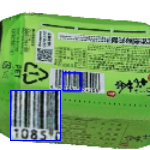}
   \vspace{1mm}
   \includegraphics[width=0.48\textwidth]{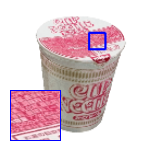}
   \includegraphics[width=0.48\textwidth]{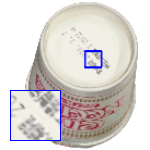}
   \vspace{1mm}
   \includegraphics[width=0.48\textwidth]{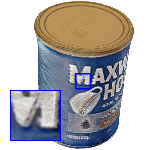}
   \includegraphics[width=0.48\textwidth]{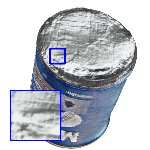}
   \vspace{1mm}
   \includegraphics[width=0.48\textwidth]{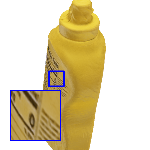}
   \includegraphics[width=0.48\textwidth]{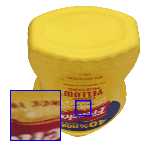}
   \vspace{1mm}
   \includegraphics[width=0.48\textwidth]{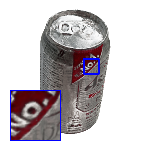}
   \includegraphics[width=0.48\textwidth]{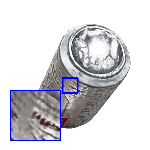}
   \vspace{1mm}
   \includegraphics[width=0.48\textwidth]{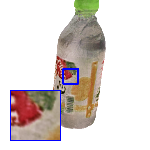}
   \includegraphics[width=0.48\textwidth]{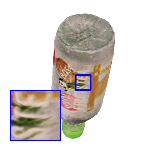}
   \caption{NeuralMeshing}
   \end{subfigure}
   \begin{subfigure}[T]{0.20\textwidth}
   \includegraphics[width=0.48\textwidth]{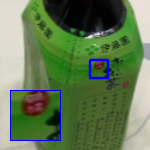}
   \includegraphics[width=0.48\textwidth]{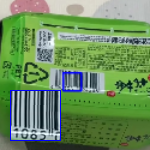}
   \vspace{1mm}
   \includegraphics[width=0.48\textwidth]{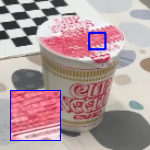}
   \includegraphics[width=0.48\textwidth]{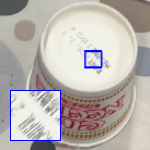}
   \vspace{1mm}
   \includegraphics[width=0.48\textwidth]{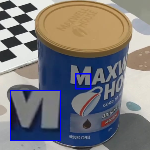}
   \includegraphics[width=0.48\textwidth]{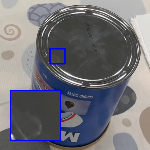}
   \vspace{1mm}
   \includegraphics[width=0.48\textwidth]{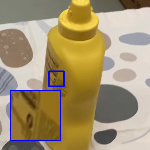}
   \includegraphics[width=0.48\textwidth]{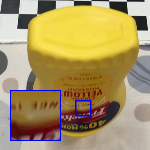}
   \vspace{1mm}
   \includegraphics[width=0.48\textwidth]{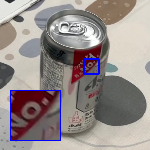}
   \includegraphics[width=0.48\textwidth]{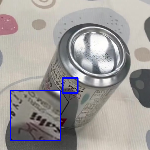}
   \vspace{1mm}
   \includegraphics[width=0.48\textwidth]{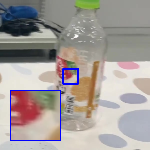}
   \includegraphics[width=0.48\textwidth]{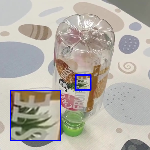}
   \caption{Ground Truth}
   \end{subfigure}
  \caption{Qualitative results}
  \label{fig:qualitative_results}
\end{figure*}

%
We recorded videos of various objects and extracted their meshes using an Apple iPhone SE.
A frame rate of 4 frames per second was used.
A checkerboard was used for automated scale estimation.
Processing was done on a PC equipped with Intel Xeon Silver 4110, 48 GB of RAM and a NVIDIA RTX A6000 GPU.
Our pipeline uses OpenCV~\cite{kaehler2014learning} and Open3D~\cite{zhou2018a} in various steps.

We report the quality of the exported meshes for five objects: A bottle of green tea (GreenTea), a cup of instant noodles (Noodles), a can of alcohol free beer (BeerCan), a bottle of mustard (Mustard) and a can of preground coffee (Coffee).
We compare methods for extracting the final mesh, either using instant-ngp, NeuS2 or our implementation with opposite surface normal sampling.
Error is calculated by rerendering the extracted meshes and comparing with the original video frames, for 20 randomly selected frames, excluding frames in which the object was significantly out of frame.
We apply the following metrics: Pixel rendering mean absolute error (MAE, lower is better), root mean square error (RMSE, lower is better) and Peak Signal to Noise Ratio (PSNR, higher is better).
Table~\ref{tab:quantitative_results} shows the results of this experiment.
Meshes extracted using NeuralMeshing have significantly higher similarity to the video frames, except for \emph{Beer Can} in which the result was similar to \emph{NeuS2}.

We report the time required for generating each mesh for the five objects above plus one additional object based on the video durations and processing time in Table~\ref{fig:timings}.
The data was collected while the operator was allowed to multitask, an operator focusing on only the mesh generation process would be able to process the meshes faster.
Processing time was always within an hour, 34 minutes on average.

\begin{table}
  \caption{Video duration and processing time (from when the video data was copied to the final mesh being exported).}
  \begin{tabular}{ ccc }
  \toprule
    Object & Video durations (seconds) & Processing time (minutes) \\
  \midrule
    GreenTea & 39, 37, 40 & 59 \\
    Noodles & 36, 36 & 28 \\
    Coffee & 34, 34 & 26 \\
    Mustard & 39, 39 & 26 \\
    BeerCan & 35, 37 & 25 \\
    FruitTea & 50, 51 & 41 \\
  \bottomrule
  \end{tabular}
  \label{fig:timings}
\end{table}

For qualitative evaluation, we include an extra item that would be difficult to quantitatively evaluate due to the lack of ground truth data, namely a plastic bottle with an almost completely transparent label (FruitTea).
Fig.~\ref{fig:qualitative_results} shows qualitative results of our method applied to the six objects.
For each object we show the rendered mesh from a viewpoint matching a frame in two different videos used to generate the mesh.
It can be seen that our method (\emph{NeuralMeshing}) performs extremely well on three out of five objects that have limited to no reflectiveness (GreenTea, Noodles, Mustard).
On reflective objects such as Coffee (the bottom is reflective) and BeerCan, performance drops significantly for all methods, including NeuralMeshing.
On FruitTea, the performance of NeuralMeshing and NeuS2 are similar, though both signifantly outperform NeRF.
FruitTea specifically shows the benefit of using neural fields based techniques for extracting meshes, as we can reasonably estimate the geometry of the object with all methods, even though automatically estimating transparency is still an open issue.

\section{DISCUSSION AND CONCLUSION}
One limitation of our system is that processing is done offline.
It takes up to an hour to generate a mesh for an object, mostly due to training and retraining neural fields at various steps in the process.
In the future we will try to reduce the processing time.

We also noticed only limited success with processing reflective or transparent objects, with errors occuring in the generated object geometry and texture.
To improve these areas in the future, we plan to explore methods to preprocess the data to generate depth~\cite{erich2023fakingdepth}, normals, albedo and roughness~\cite{Liang_2025_CVPR} and methods for detecting the opacity of pixels~\cite{erich2025,yao2024vitmatte}.
Dedicated techniques for metallic object mesh extraction using neural representations exist~\cite{10.1145/3592134}, however these typically require a few hours of training time and are thus infeasible to be integrated into our approach.
In our quantitative evaluation we excluded the FruitTea object.
Ground truth data for largely transparent objects cannot easily be obtained as transparent surfaces transmit the background color of the surface behind the transparent surface, which depends on the viewpoint, while triangular mesh vertex color is viewpoint independent.
Even if we could encode a viewpoint dependent background color into the final mesh, it could be argued that it is more desirable to have a neutral color and a per vertex transparency value.
Transparency estimation can be evaluated by rerendering the object in the same environment but with the physical object removed, as we did in an earlier study~\cite{erich2025}, however such evaluation requires a more controlled environment (e.g.\ camera pose should be static).

We tested various cameras for recording the datasets.
Cheap USB cameras such as Sanwa CMS-V45S struggle with high resolution image capture due to image warping.
More expensive USB cameras such as Logitech BRIO might still struggle due to autofocus causing some images to become blurry.
Best results were obtained with the Apple iPhone SE's camera and with cameras embedded in RealSense sensors.

Even though we used a checkerboard for automated scale estimation, we did not use it to input intrinsics in the Structure-from-Motion algorithm or for generating camera poses.
The checkerboard is only visible in a subset of the frames, thus it is not possible to detect all the camera poses using it.
Other markers such as AR markers or a ChArUco board might be easier to embed in each frame, which would allow skipping the SfM step.

In this paper we adopt NeuS2 for meshing, however there are other techniques for generating meshes from volumetric data such as nerf2mesh~\cite{tang2023g} and SuGaR~\cite{guedon2024a}.
In the future we will investigate integrating such alternative mesh generation components into our system.

Our pipeline heavily relies on COLMAP~\cite{schoenberger2016mvs, schoenberger2016sfm} and GLOMAP~\cite{10.1007/978-3-031-73661-2_4} for estimating the camera poses.
These methods use a multi-stage computational process to convert images into camera poses.
Recently various new machine learned methods for camera pose estimation have appeared, such as DUSt3R~\cite{dust3r_cvpr24} and VGGT~\cite{wang2025vggt}.
DUSt3R offers similar performance as COLMAP and GLOMAP, but can be used with significantly fewer cameras.
However, as our end goal is mesh generation, having fewer frames negatively affects other steps such as neural field construction.
VGGT can outperform COLMAP and GLOMAP in certain scenarios, so in the future we will try to embed it into our pipeline to evaluate its effectiveness for our application.
MUSt3R~\cite{Cabon_2025_CVPR} extends DUSt3R to enable rapid camera pose estimation for a large number of views, and is thus also interesting to consider as an alterative to VGGT.
\section*{ACKNOWLEDGMENT}
This work was supported by JST [Moonshot R\&D][Grant Number JPMJMS2031].

\flushend
\bibliographystyle{IEEEtran}
\bibliography{IEEEabrv,SII2026}

%

\end{document}